\documentclass {article}
\pdfoutput=1
\usepackage{spconf}
\usepackage{listings}
\usepackage{cite}
\usepackage{url}
\usepackage[linesnumbered,ruled,vlined]{algorithm2e}
\usepackage{float}
\usepackage{graphicx}
\usepackage[T1]{fontenc}	
\usepackage{amsmath}		
\usepackage{amssymb}		
\restylefloat{figure}

\newcommand{\R}{{\mathbb R}}

\newcommand{\be}{\begin{equation}}
\newcommand{\ee}{\end{equation}}
\newcommand{\ba}{\begin{array}}
\newcommand{\ea}{\end{array}}
\newcommand{\baa}{\left[\begin{array}}
\newcommand{\eaa}{\end{array}\right]}

\newcommand{\beqa}{\begin{eqnarray}}
\newcommand{\eeqa}{\end{eqnarray}}
\newcommand{\bt}{\begin{tabular}}
\newcommand{\et}{\end{tabular}}

\newcommand{\bi}{\begin{itemize}}

\newcommand{\ei}{\end{itemize}}
\newcommand{\bc}{\begin{center}}
\newcommand{\ec}{\end{center}}

\name{Paul Irofti \thanks{This work was supported by the Romanian National Authority
	   for Scientific Research, CNCS - UEFISCDI, project number PN-II-ID-PCE-2011-3-0400 and by the Sectoral Operational Programme
       Human Resources Development 2007-2013 of the Ministry of European
       Funds through the Financial Agreement POSDRU/159/1.5/S/132395.
	   E-mail: paul@irofti.net.}
}
\title{Efficient GPU Implementation for Single Block Orthogonal \\
	Dictionary Learning}
\address{
	  Department of Automatic Control and Computers \\
      University Politehnica of Bucharest \\
      313 Spl. Independen\c{t}ei, 060042 Bucharest, Romania
}

\begin {document}
\maketitle

\begin{abstract}
Dictionary training for sparse representations involves dealing with large
chunks of data and complex algorithms that determine time consuming
implementations. SBO is an iterative dictionary learning algorithm based on
constructing unions of orthonormal bases via singular value decomposition, that
represents each data item through a single best fit orthobase. In this paper we
present a GPGPU approach of implementing SBO in OpenCL. We provide a lock-free
solution that ensures full-occupancy of the GPU by following the map-reduce
model for the sparse-coding stage and by making use of the Partitioned Global
Address Space (PGAS) model for developing parallel dictionary updates. The
resulting implementation achieves a favourable trade-off between algorithm
complexity and data representation quality compared to PAK-SVD which is the
standard overcomplete dictionary learning approach. We present and discuss
numerical results showing a significant acceleration of the execution time for
the dictionary learning process.
\end{abstract}

\section{Introduction}

The sparse representations field is the basis for a
wide range of very effective signal processing techniques with
numerous applications for, but not limited to, audio and image
processing.

Such applications fall naturally within the realm of parallel GPU-computing due to the data size and the way the algorithms process it.
When it comes to implementations, recent years have shown a tendency towards
OpenCL mainly because of its portable nature and wide industry support.

In this paper, we approach the problem of training dictionaries for sparse
representations by learning from a representative data set.
The goal is that given a set of signals $Y \in \R^{p \times m}$
and a sparsity level $s_0$
to find a dictionary $D \in \R^{p \times n}$
that minimizes the Frobenius norm of the approximation error
\be
E = Y - DX
\label{eq:frob}
\ee
where $X \in \R^{n \times m}$ is the associated sparse representations matrix
that uses $s_0$ columns (or atoms) from $D$
for sparse coding each column (or data-item) from $Y$.

This is a difficult problem because both the dictionary $D$ and the sparse
representations $X$ are unknown and so existing solutions (K-SVD\cite{AEB06},
AK-SVD\cite{RZE08}, UONB\cite{LGBB05}, SBO\cite{RD13}) approach this
as an optimization problem solved via alternative iterations.
We express this as a minimization of the Frobenius norm from (\ref{eq:frob})
with an $l_0$-norm sparsity constraint:
\be
\begin{aligned}
\label{eq:optimize}
& \underset{D,X}{\text{minimize}}
&& \|Y - DX\|^2_F \\
& \text{subject to}
&& \|x_i\|_0 \leq s_0, \; \forall i
\end{aligned}
\ee

More specific,
first the dictionary is fixed and the sparse representations are found
by applying an algorithm such as OMP\cite{PRK93}
and then, keeping the representations fixed, the dictionary is refined
by updating or expanding its content.

While the generic dictionary learning problem doesn't impose any specific
structure on the dictionary $D$, some methods\cite{LGBB05}\cite{RD13}
build the dictionary
as a union of smaller blocks consisting of ortonormal bases (ONBs)
that transform the optimization problem into:
\be
\begin{aligned}
\label{eq:block_optimize}
& \underset{D,X}{\text{minimize}}
&& \|Y - [Q_1 Q_2 \dots Q_K ]X\|^2_F \\
& \text{subject to}
&& \|x_i\|_0 \leq s_0, \; \forall i \\
&&& Q_j^TQ_j = I_p, \; 1 \leq j \leq K
\end{aligned}
\ee

where the union of $K$ ONBs denoted
$Q_j \in \R^{p \times p}$, with $j = 1 \dots K$, represents the dictionary $D$.

The union of orthonormal basis algorithm (UONB) and the single block orthogonal
(SBO) algorithm enforce this structure on the dictionary by using
singular value decomposition (SVD) to create each orthonormal block.
The difference between the two is that for representing a single data item
the former uses atoms selected via OMP from all bases,
while the later uses atoms from a single orthoblock.
Because of its representation strategy, SBO uses more dictionary blocks than
UONB but also executes faster while maintaining the same representation error.

We are interested in parallelizing SBO because
it brings data-decoupling through its single block representation system
and also because it doesn't depend on OMP
which raised hard full GPU occupancy problems, even when applying the PGAS
method, due to its high memory footprint\cite{ID14}.

\section{The SBO algorithm}
SBO builds the dictionary as a union of orthoblocks
and forces each data-item from $Y$ to use a single block $Q_j$ for its sparse
representation $x$ such that $y \approx Q_j x$.
The representation of $x$ results from computing the product $x = Q_j^T y$
and then hard-thresholding the $s_0$ highest absolute value entries.
The orthonormal base $Q_j$ is picked by
computing the energy of the representation coefficients from $x$
and selecting the orthobase where the energy is highest:

\be
j = \underset{i = 1\dots K}{\text{argmax}} \sum_{s=1}^{s_0}{|Q_i^T y|}
\label{eq:alloc}
\ee

Following this method, each data-item from $Y$ is represented by
a single orthobase in a process that we'll call representation.

The alternative optimization iterations for performing dictionary learning
on a single orthonormal base is presented in algorithm \ref{algo:1onb}.

\begin{algorithm}
\label{algo:1onb}
\SetKwComment{Comment}{}{}

\BlankLine
\KwData{\quad signals set $Y$,
	initial dictionary $Q_0$, \\
	\hspace{12mm} target sparsity $s_0$,
	number of rounds $R$
}
\KwResult{\hspace{1mm} trained dictionary $Q$,
	sparse coding $X$
}
\BlankLine

$ Q = Q_0 $ \\
\For{$r \leftarrow 1$ \KwTo $R$}{
	$X = Q^T Y$ \\
	$X(:,j) =$ SELECT$(X(:,j), s_0)$ \\
	$P = Y X^T$ \\
	$U \Sigma V^T =$ SVD$(P)$ \\
	$Q = U V^T$
}
\caption{1ONB}
\end{algorithm}

By keeping a fixed dictionary $Q$, step 3 computes the new
representations $X$ and step 4 performs hard-thresholding through
partial sorting to select the largest $s_0$ values on each
column. Using the new matrix $X$, the dictionary is refined (step 7)
by using the product of the resulting orthonormal matrices from the SVD
computation in step 6.

The results from \cite{LGBB05} show that, with a good initialization (step 1),
good results can be reached by just a few iterations ($R < 5$ in step 2).
Also, a good starting point when creating a new orthoblock
is to use the left-hand side orthonormal matrix of the SVD decomposition
of the given data set:

\be
Y = U\Sigma V \rightarrow Q_0 = U
\label{eq:base-init}
\ee

Based on the above, SBO is described in algorithm \ref{algo:sbo}.

\begin{algorithm}
\label{algo:sbo}
\SetKwComment{Comment}{}{}

\BlankLine
\Comment{Initialization}
\BlankLine

Iteratively train $K_0$ orthonormal blocks by randomly selecting $P_0$ signals
from $Y$ and applying 1ONB $R$ times:
$D = [ \ba{ccc} Q_1 & \dots & Q_{K_0} \ea ]$\\

Represent each data-item with only one of the previously computed ONBs
following (\ref{eq:alloc})\\

\BlankLine
\Comment{Iterations}
\BlankLine

	Construct the set of the worst $W$ represented data items and
	train a new orthobase with this set. Add the new base to the existing
	union of ONBs.\\

	Represent each data item with one ONB\\

	Train each orthobase over its new data set\\

	Check stopping criterion\\
\caption{SBO}
\end{algorithm}

The method is split in two parts: the initialization phase and the dictionary
learning iterations.

The initialization phase builds a small start-up dictionary consisting of $K_0$
orthobases each trained with $P_0$ sized signal chunks that are used
by 1ONB to initialize and train a new orthobase (step 1).
The resulting dictionary is used by step 2 to perform data item representation
which leads to an initial sparse representation set.

The training iterations start by building a new orthobase for the worst
$W$ represented signals using algorithm \ref{algo:1onb} and expanding
the dictionary to include the new ONB (step 3).
Given that the dictionary has changed, a new data-item representation is needed
and with that step 4 computes a new set of sparse representations.
Step 5 refines the dictionary $D$ by applying 1ONB
on each orthobase over its newly associated data set.
The learning process is stopped by either reaching a given target error
or the permitted maximum number of orthonormals.

\section{Parallel SBO with OpenCL}

In this section we will go through the main points behind our parallel version,
then give some details on the OpenCL specifics.

\subsection{Parallel representations}
\label{subsec:onb-alloc}

The sparse representations are completely independent and so their computation
is done in parallel by applying (\ref{eq:alloc}) on each data-item.
More specific, for each signal from $Y$ we compute the representations
with every available orthoblock and pick the one that has the highest energy.
As shown in \cite{RD13}, computing the energy is enough.

This task fits naturally on the map-reduce model.
We map the data in signal-orthobase pairs that produce the energy
of the resulting sparse representation.
Each pair computes the representation with the current dictionary block $j$
($x = Q_j^T y$), does a hard-threshold on the largest $s_0$ items in
absolute value, and outputs the energy $E$ of the resulting sparse coding.
Parallelization is done in bulk by performing the above for all ONBs at once
in groups of $\tilde m$ signals.
The result is that each data item has an associated energy list of its
representation with each block from the dictionary.
We reduce the list, for each signal in $Y$, to the element with the largest
energy leading to the choice of a single representation block.

\begin{figure}
\bc
\framebox{\includegraphics[width=0.90\linewidth]{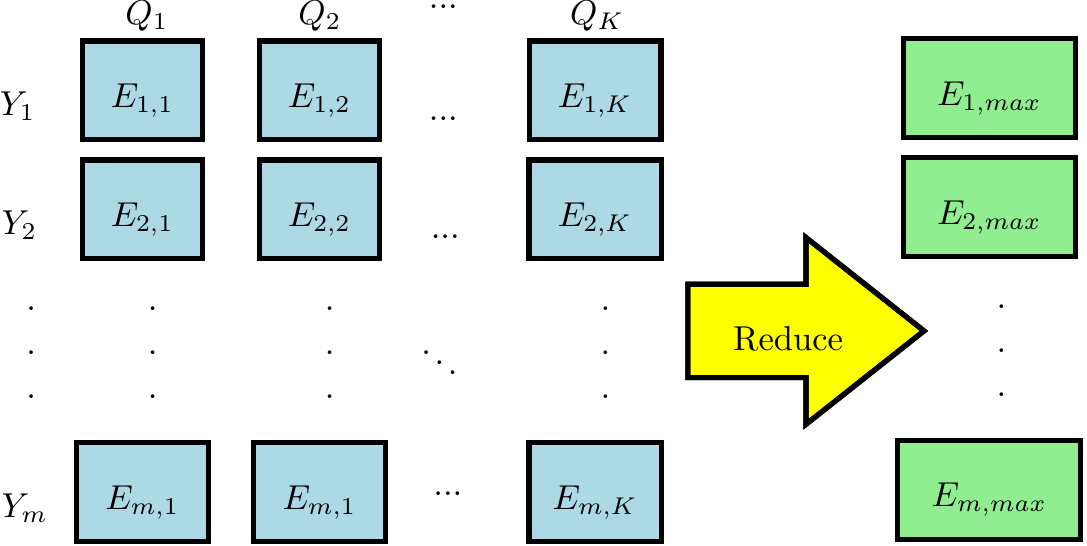}}
\ec
\caption{MapReduce for $\tilde{m}=1$ and $K$ orthobases}
\label{fig:mapreduce}
\end{figure}

\subsection{Parallel dictionary training}
\label{subsec:onb-train}

Dictionary learning is performed by the operations of 1ONB
described in algorithm \ref{algo:1onb}.
SBO makes use of 1ONB in three different contexts:
once during the initialization phase (step 1),
and twice during the training iterations while
learning a new dictionary for the $W$ worst represented signals (step 3)
and while training the existing dictionary over its new data set (step 5).

\begin{figure}
\bc
\framebox{\includegraphics[width=0.90\linewidth]{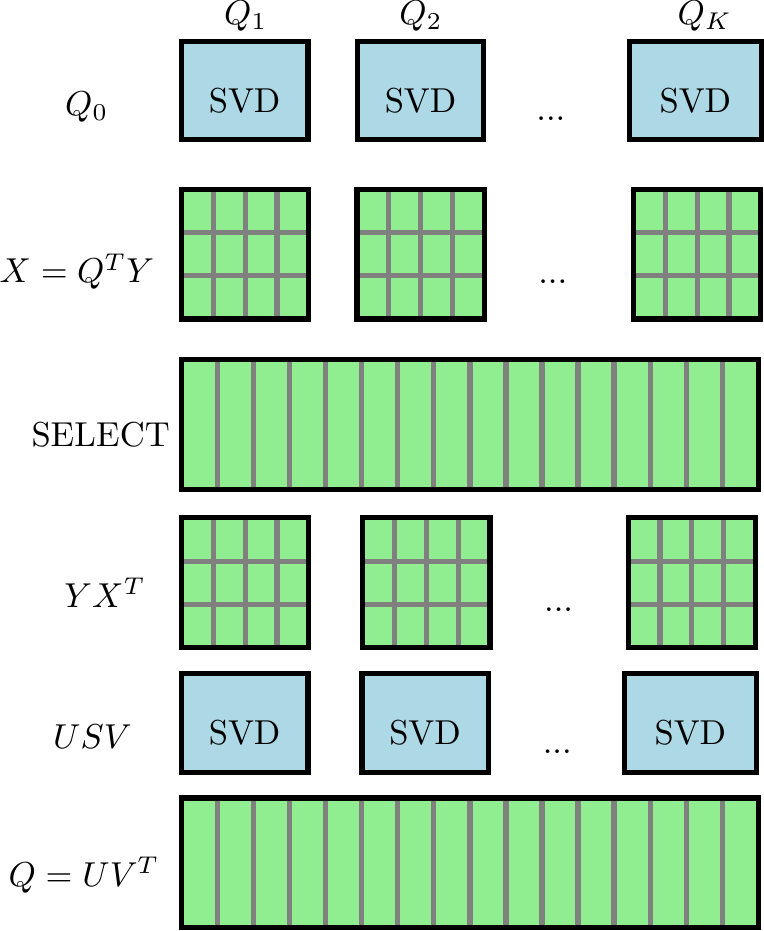}}
\ec
\caption{The parallel execution of 1ONB for $R=1$ rounds and $K$ orthobases.
Each block represents a task and each sub-block depicts a thread of execution
within that task.}
\label{fig:1onb}
\end{figure}

Due to the decoupled nature of the data,
we add parallelism at the dictionary level
(each orthoblock is initialized and trained in parallel)
and we also further parallelize the steps of each orthoblock training
instance (see figure \ref{fig:1onb}).
This approach allows us to execute the sequential operations inside 1ONB
(mainly the SVD routines) in parallel for each dictionary block.

If an initial orthonormal basis is not supplied,
we generate a new basis by using the singular value decomposition
as described in (\ref{eq:base-init}). This, along with the other SVD operation
from step 6 are executed in parallel for each dictionary block.
The alternative optimization iterations (steps 3--6) train the orthonormal
dictionary $Q$ such that $\| Y - QX \|_F$ is minimized or reduced.
First, keeping a fixed dictionary, the sparse representations are
computed in step 4.
Since this is done via matrix multiplication of large dimensions
it can be easily parallelized through the classic concurrent sub-block
multiplication routines.
The target sparsity is obtained by hard-thresholding the largest $s_0$
absolute value entries (step 4).
We compute the thresholding in parallel for groups of $\tilde m$ signals by
evenly partitioning the global address space for each thread of execution.
Second, using the new matrix $X$, we update the dictionary
via the SVD decompositon (step 6) of $YX^T$ from step 5 by using the resulting
orthonormal matrices $U$ and $V$ (step 7).
We perform the $YX^T$ matrix multiplication and the decomposition in parallel
just as we did before.
Step 7 represents a matrix multiplication of relatively small dimensions
($p \times p$) for which analysis showed that it is better to employ a PGAS
strategy so that each thread performs a few corresponding vector-matrix
operations resulting in a simultaneous update of all orthobases.

\subsection{OpenCL implementation details}
\label{subsec:opencl}

The OpenCL platform allows us to execute small functions (kernels) in
parallel on a chosen number of processing elements (PE), or work-items,
within the compute units located on the OpenCL device\cite{opencl}.
These PEs are organized in an n-dimensional space that can be
set up in different ways for each kernel.
The n-dimensional space is split into local work-groups,
corresponding to compute units; PEs in a work-group can better
share common resources.
For example, in 2D, we can denote the n-dimensional range definition as
NDR($\langle x_g,y_g \rangle$, $\langle x_l, y_l \rangle$).
There are $x_g \times y_g$ PEs, organized on work-groups of size $x_l \times y_l$,
running the same kernel.

{\em Matrix multiplication.}
Steps 3 and 5 from the 1ONB algorithm were implemented using the BLAS library
for OpenCL from AMD.
The AMD kernels follow the classic GEMM BLAS model.
Input matrices and the result are stored in global memory.
The operation first creates matrix sub-groups and then does block-based
full-matrix multiplication on them.
While the AMD implementation doesn't take full advantage of the hardware
underneath, it's fast enough for our use-case.
We compensate it poor occupancy of the GPU resources
(profiling our simulations with AMD's CodeXL showed
33.3\% for the sub-grouping and 25\% for the block multiplication)
by scheduling as many GEMM operations at the same time as there are orthobasis
(SBO step 1 and step 5).

{\em Representation.}
Given $k$ orthoblocks, all the operations required for finding the
best dictionary block for the sparse representation
of each data item from the signal set,
SBO step 2 and 4,
were packed and implemented by a single OpenCL kernel
following the optimization problem (\ref{eq:alloc}).

The input matrices as well as the resulting orthobase representation index
of each signal and its energy are kept in global memory.
We can keep the actual sparse representations in private memory
because only the energy and base representation indices are needed by SBO.
During representation, the sparse signal storage is accessed multiple
times for each orthobase in order to compute $x=Q^T y$.
Keeping the memory private gains us low latency times at the
expense of an increased number of vector general purpose registers used which,
in turn, leads to a lower occupancy level.
Our numeric experiments showed that lower latency outbids by far a
partitioned global memory, full-occupancy version of the kernel.

We designed the representation kernel following the map-reduce paradigm.
We map each work-item to a signal-orthoblock couple.
Each processing element is in charge of sparse coding
and computing the resulting energy of a few $\tilde{m}$ signals
using a single orthobase.
The energy is saved in a matrix in local memory at the signal-orthobase
coordinates corresponding to the work-item's position in the work-group.
We keep 2-dimensional work-groups with orthobases in the first dimension and
signals on the second as depicted on the left side of
figure \ref{fig:mapreduce}.
And so we split the signal set in $\tilde m$ sized chunks representing the
number of work-groups scheduled for processing on the compute-units,
corresponding to an
$NDR(\langle k, m\rangle, \langle k, \tilde m\rangle)$ splitting.
The reduction on the columns of the energy matrix is performed
by each work-item with ID 0 in the orthobase dimension
(see the right-side of figure \ref{fig:mapreduce}).
Even though this approach leaves most of the work-items idling when reducing,
the overhead of doing map-reduce in the same kernel
(opposed to doing it in two separate ones)
is insignificant in this case.

{\em Dictionary training.}
The dictionary update process, SBO step 1 and step 5,
was split into parts and implemented by multiple OpenCL kernels.
We keep the input matrices for the dictionary and the signal set in global
memory as well as the resulting sparse representations.
The dictionary bases are modified in-place.

Before starting the dictionary training phase in SBO's step 5,
we group the signals in blocks based on the dictionary-base used for their
representations.
This speeds-up the training process by using coalesced memory
in SBO's parallel implementation.
We first build a list of signals for each base $Q$
and then we walk it contiguously copying
the signals using $Q$ overwriting the matrix $Y$.
This is a cheap operation that brings a big performance boost
by helping data access times of the execution threads.
Copying proved to be up to $1000\times$ more effective
by mapping the signal matrix in host memory and using $memcpy$
than plainly using $clEnqueueCopyBuffer$.

For the implementation of algorithm \ref{algo:1onb} we decided to use
a Numerical Recipes based implementation of the SVD algorithm.
We execute it in parallel through an OpenCL kernel for each orthoblock
on the GPU with an $NDR(\langle k \rangle, \langle 1 \rangle)$ splitting.
The matrix multiplications (steps 3 and 5), as discussed earlier,
are processed by the BLAS kernels from AMD.

The operations for partial selection (step 4 in 1ONB)
were packed and implemented as a separate OpenCL kernel.
The sparse signal set is kept in global memory and each work-item
is in charge of doing SELECT on $\tilde m$ signals.
Numerical experiments on our hardware pointed out that a splitting of
$NDR(\langle m \rangle, \langle \tilde m=256 \rangle)$
gives the best performance results while keeping full GPU occupancy.

Due to the small dimensions $p$ of the block dictionaries, using the BLAS
library from AMD for processing step 7 of 1ONB for each orthobase
didn't cover the IO costs.
For that, we implemented a custom matrix multiplication kernel that
performs the operation in parallel for the entire dictionary.
And so, each work-group is in charge of computing the updated orthobase
corresponding to its group-id, resulting in an
$NDR(\langle k \times \tilde m \rangle, \langle \tilde m \rangle)$
splitting.
Work-items within a work-group are performing vectorized
vector-matrix multiplication
that calculate the lines of the new orthobase corresponding to their local-id.
Given that $Q \in \R^{p \times p}$, the number of lines each work-item
has to compute is given by the ratio of $p / \tilde m$.
For $p$ dimensioned $k$ orthobases we found that a subunitary ratio of the form
$NDR(\langle k \times \tilde m \rangle, \langle \tilde m=p\times8 \rangle)$
gives full occupancy on our GPU.

Updating the energy of the newly created sparse representations
(needed in step 3 of the next SBO iteration for building the worst
represented signals set $W$)
is implemented following the PGAS model by another OpenCL kernel.
The representation matrix and the associated energy set are kept in
global memory.
Each work-item independently computes the energy for $m / \tilde m$
signals with no work-group cooperation resulting in an
$NDR(\langle \tilde m \rangle, \langle any \rangle)$
split.
We found that full-occupancy is reached on our hardware by using the
$NDR(\langle K \times l\rangle, \langle l=192 \rangle)$
partitioning, where $K$ is the maximum allowed number of orthobases.

\section{Results and Performance}
We used colored and gray scale bitmap images for the training signals, taken
from the USC-SIPI \cite{sipi} image database (e.g. barb, lena, boat, etc.).
The images were normalized and split into random $8 \times 8$ blocks

As a rule, we chose the dimensions as powers of two because this way the
data objects and the work-loads are easier divided and mapped across the NDRs
without the need for padding.

We tested our OpenCL implementation of SBO on an
ATI FirePro V8800 (FireGL V) card from AMD,
running at a maximum clock frequency of 825MHz,
having 1600 streaming processors,
2GB global memory and 32KB local memory.
Also, the CPU tests for our C implementation were made on an
Intel i7-3930K CPU
running at a maximum clock frequency of 3.2GHz.

Tables \ref{tab:paksvd-rmse} and \ref{tab:sbo-rmse} depict the differences
in final representation error, the total time spent on dictionary learning
($t_{learn}$) and the time it takes to represent the data set with the final
dictionary ($t_{rep}$).
We vary the total number of SBO orthoblocks $K=\{8,16,32,64\}$ and
compare with PAK-SVD instances running with
a dictionary of $n=\{64,96,128,256\}$ atoms and $K=100$ iterations
using full parallelization during the atoms update phase ($n = \tilde{n}$)
for which the numerical simulations in \cite{ID14} gave the best
representation error and the fastest execution times.
The RMSE is $\|Y - DX\|_F / \sqrt{pm}$.
While PAK-SVD can produce a slightly better error than SBO,
the time difference is significant with SBO being up to 203.8 times faster
than PAK-SVD at dictionary learning and 1068.4 times faster at producing sparse
representations.
Even though SBO's dictionary size is larger, the total memory
footprint is smaller than PAK-SVD because of OMP's high memory requirements.

\begin{table}
\bc \bt{ |c |  c | c | c | c | c | }
\hline
$n$ & 64 & 96 & 128 & 160 & 256 \\
\hline
$t_{learn}(s)$ & 366.8 & 396.7 & 416.5 & 438.4 & 642.4 \\
\hline
$t_{rep}(s)$ & 0.3467 & 0.3753 & 0.8207 & 0.5889 & 2.2436 \\
\hline
RMSE & 0.0271 & 0.0246 & 0.0242 & 0.0230 & 0.0216 \\
\hline
\et \ec
\caption{PAK-SVD performance for $m=32768$, $p=64$, $s_0=8$
	with $\tilde n = n$ and $K=100$.}
\label{tab:paksvd-rmse}
\end{table}

\begin{table}
\bc \bt{ |c |  c | c | c | c | c | }
\hline
$K$ &8 &16 &24 &32 & 64 \\
\hline
$t_{learn}(s)$ & 1.8 &6.7 &12.3 &20.9 &85.4 \\
\hline
$t_{rep}(s)$&0.0020 & 0.0021 & 0.0022 & 0.0021 & 0.0021 \\
\hline
RMSE & 0.0268 &0.0245 &0.0240 &0.0238 &0.0235 \\
\hline
\et \ec
\caption{Parallel SBO performance for $m=32768$, $p=64$, $s_0=8$
	with $K_0=5$ and $R=6$}
\label{tab:sbo-rmse}
\end{table}

\begin{figure}[H]
\bc \framebox{\includegraphics[width=0.90\linewidth]{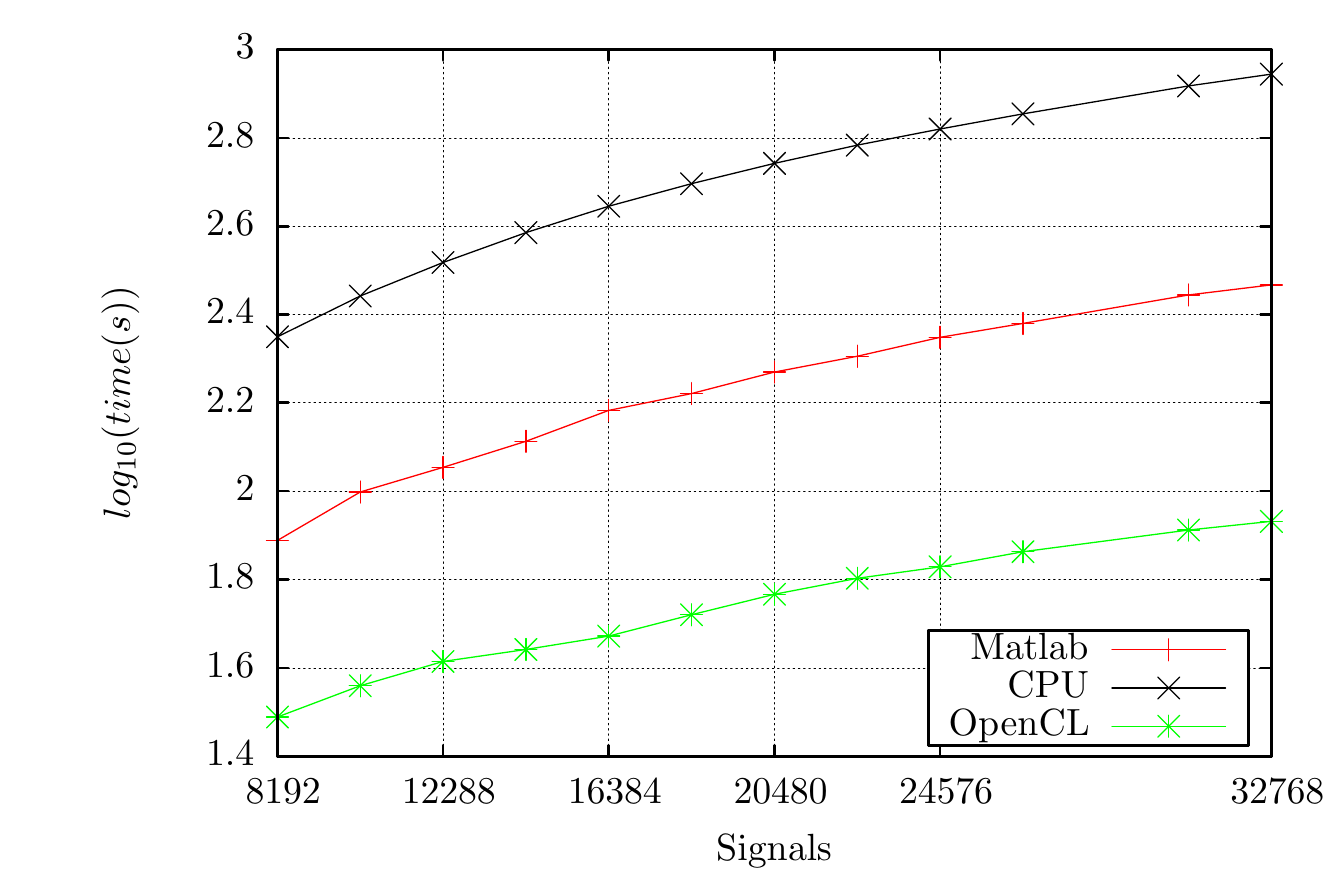}} \ec
\caption{Execution times for $K=64$, $s_0=8$, $p=64$.}
\label{fig:perf-signals}
\end{figure}

Turning our focus towards different SBO implementations,
we see in figure \ref{fig:perf-signals}
that the OpenCL implementation gives better results than the Matlab and C
counterparts.
Keeping a fixed number of orthonormal bases $K=64$ and representing signal sets
from as low as $m=8192$ up to $m=32768$, the parallel version performs
3.4 times faster than the Matlab implementation
and 10.3 times faster than the single CPU C implementation.  

\begin{figure}[H]
\bc \framebox{\includegraphics[width=0.90\linewidth]{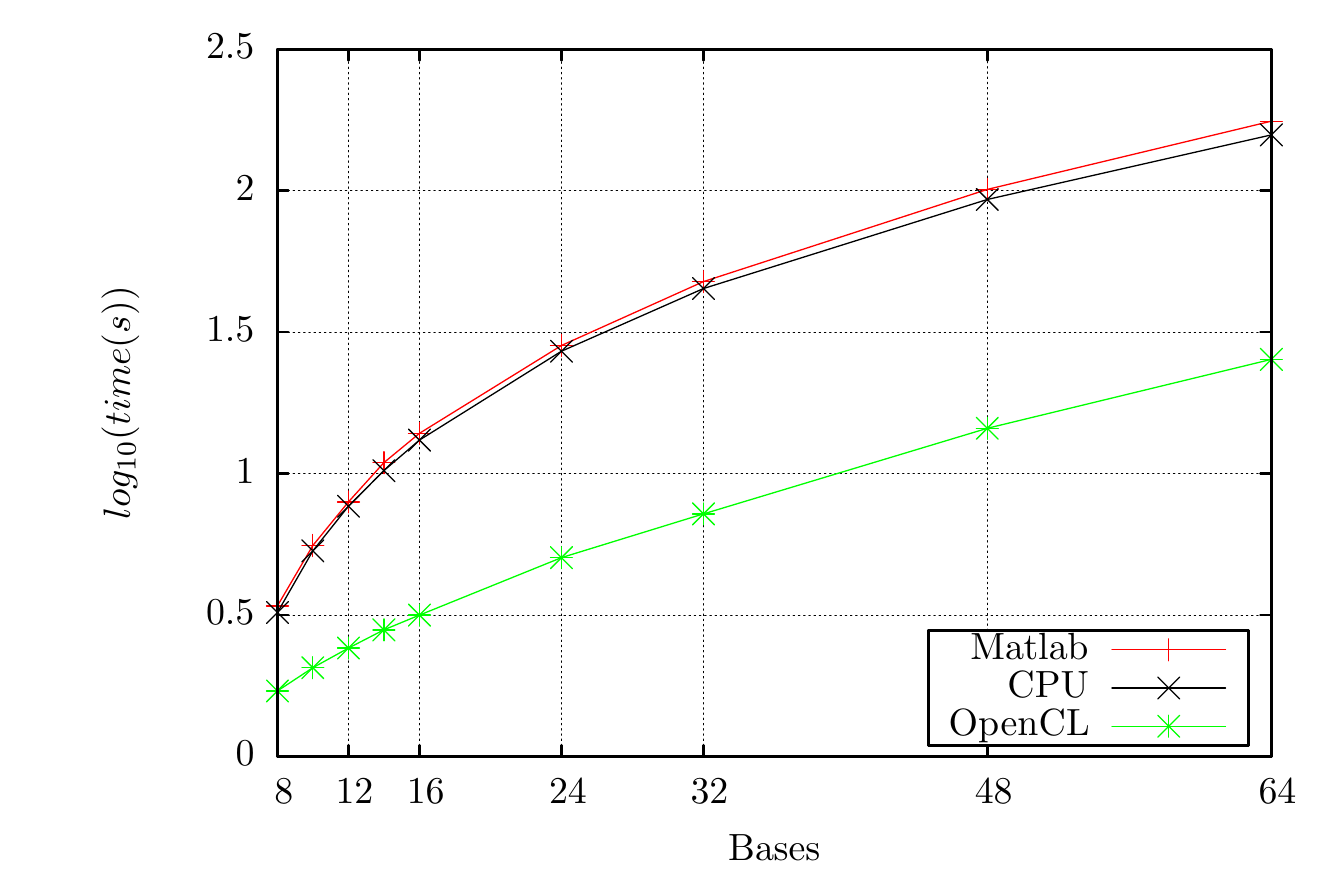}} \ec
\caption{Execution times for $m=24576$, $s_0=4$, $p=32$.}
\label{fig:perf-bases}
\end{figure}

Figure \ref{fig:perf-bases} describes the performance results with a fixed
signal set of $m=24576$ and a variable dictionary size starting from
$K=8$ orthoblocks up to $K=64$.
Again we can see that the OpenCL version performs a lot better than the
other implementations, giving speed-ups up to 7 times.

\begin{figure}
\bc \framebox{\includegraphics[width=0.90\linewidth]{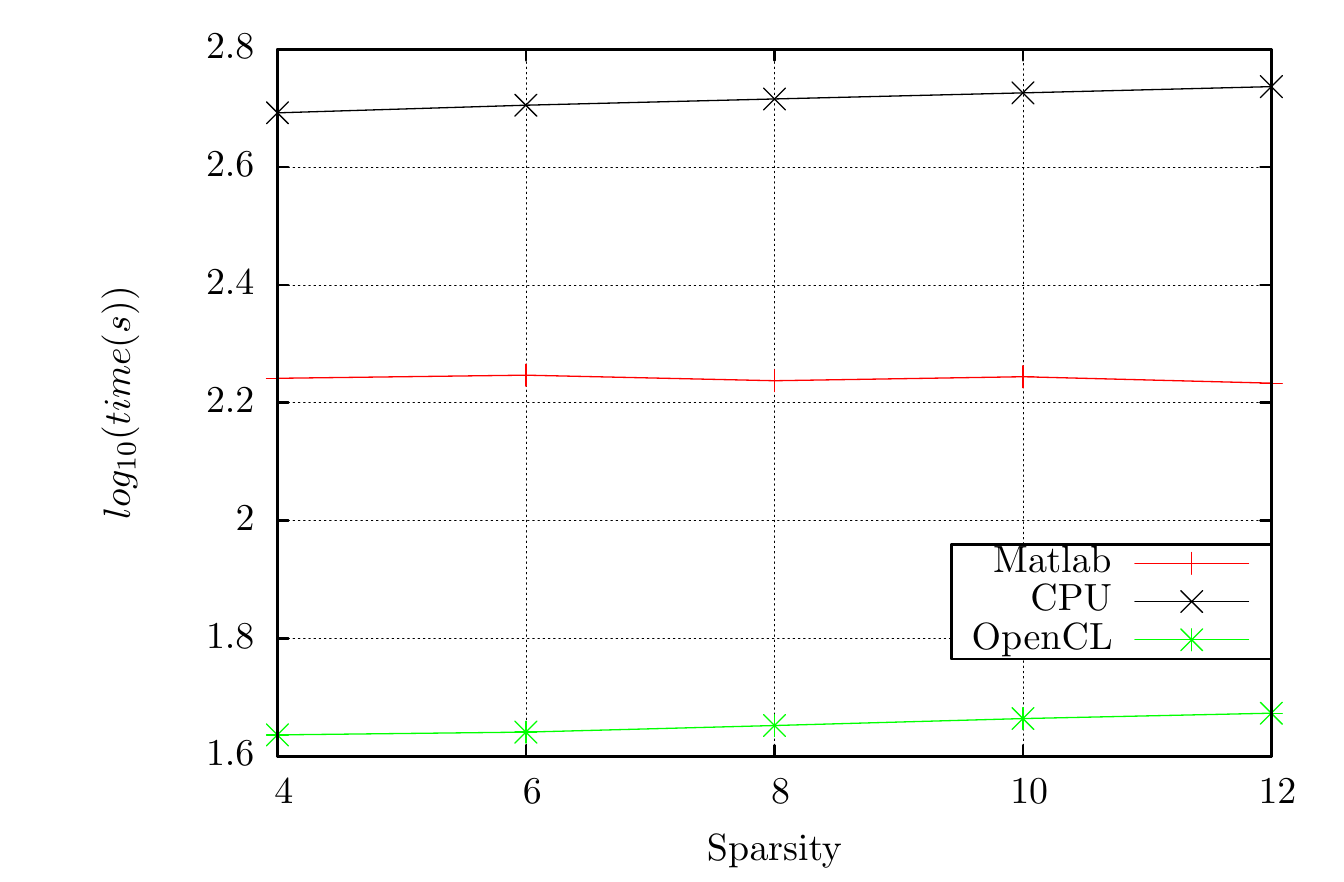}} \ec
\caption{Execution times for $m=32768$, $K=48$, $p=64$.}
\label{fig:perf-sparsity}
\end{figure}

Looking at figure \ref{fig:perf-sparsity} we see that the target sparsity $s_0$
doesn't really affect running times.
We kept a fixed signal set $m=32768$
and a fixed dictionary of $K=48$,
and we varied the sparsity from $s_0=4$ to $s_0=12$
on a fixed signal dimension of $p=64$.

\section{Conclusions}

We have proposed an efficient parallel implementation of the SBO algorithm.
Dictionary updates are performed by refining each of the orthonormal
bases concurrently.
Also, we completely parallelized, in a map-reduce manner,
the pursuit of finding the single best orthobase for representing a
given signal.
Our implementation was done in OpenCL and tested on the GPU.

Our parallel version achieves a good trade-off between algorithm
complexity and data-set approximations compared to PAK-SVD due to
the different representation approach and
the low-memory footprint of SBO's representation strategy
leading to better GPU occupancy
confirmed in our numerical results that show a speed-up of about
200 times for dictionary learning
while providing almost the same representation quality.
Despite its much larger dictionary size, SBO has a significantly
lower representation time
(simulations show about 1000 times speed improvement),
which makes it appealing for real time applications.
Also, simulations showed that the SBO OpenCL version can perform about
10 times faster on the same data than the sequential versions.

\bibliographystyle{IEEEbib}
\bibliography{sparse}

\end{document}